\documentclass{article}
\usepackage{spconf,amsmath,graphicx}
\usepackage{tikz}
\usepackage{pgfplots}
\usepackage{graphicx}
\usepackage{multirow}
\usepackage{enumerate}

\newcommand{\myfigref}[1]{Fig. \ref{#1}}
\newcommand{\mytabref}[1]{Tab. \ref{#1}}

\title{RATs-NAS: Redirection of Adjacent Trails on GCN for Neural Architecture Search}
\name{$^1$Yu-Ming Zhang, $^2$Jun-Wei Hsieh, $^1$Chun-Chieh Lee, $^1$Kuo-Chin Fan}
\address{$^1$Dep. of Computer Science and Inf. Eng., National Central University, Taoyuan, Taiwan\\
         $^2$College of AI and Green Energy, National Yang Ming Chiao Tung University, Hsinchu, Taiwan}

\begin{document}
%
\maketitle
\begin{abstract}
Various hand-designed CNN architectures have been developed, such as VGG, ResNet, DenseNet, etc., and achieve State-of-the-Art (SoTA) levels on different tasks. Neural Architecture Search (NAS) now focuses on automatically finding the best CNN architecture to handle the above tasks. However, the verification of a searched architecture is very time-consuming and makes predictor-based methods become an essential and important branch of NAS. Two commonly used techniques to build predictors are graph-convolution networks (GCN) and multilayer perceptron (MLP). In this paper, we consider the difference between GCN and MLP on adjacent operation trails and then propose the Redirected Adjacent Trails NAS (RATs-NAS) to quickly search for the desired neural network architecture. The RATs-NAS consists of two components: the Redirected Adjacent Trails GCN (RATs-GCN) and the Predictor-based Search Space Sampling (P3S) module. RATs-GCN can change trails and their strengths to search for a better neural network architecture. P3S can rapidly focus on tighter intervals of FLOPs in the search space. Based on our observations on cell-based NAS, we believe that architectures with similar FLOPs will perform similarly. Finally, the RATs-NAS consisting of RATs-GCN and P3S beats WeakNAS, Arch-Graph, and others by a significant margin on three sub-datasets of NASBench-201.
\end{abstract}


\section{Introduction}
Many Convolution Neural Networks (CNN) have been proposed and achieved great success in the past decade \cite{vgg16, resnet, densnet}. However, designing a handcrafted CNN architecture requires human intuition and experience, which takes work and time to build an optimal CNN. NAS \cite{nas} focuses on this problem and automatically searches for the best neural network architecture based on a specific strategy in the specific search space \cite{nasbench101, nasbench201}. Many methods have been proposed recently. The cell-based NAS relies on a meta-architecture to reduce the complexity of the search scale. The meta-architecture is a CNN model with pre-defined hyperparameters, such as the number of channels and stacked cells. Those stacked cells are composed of operations such as convolution, pooling, etc. Therefore, searching for a CNN architecture is equivalent to searching for a cell. However, it is time-consuming to verify a searched architecutre candidate. The predictor-based NAS method encodes an architecture with an adjacency matrix and an operation matrix to quickly predict an architecture's performance. The adjacency matrix indicates the adjacent trails of operations in a cell, and the operation matrix indicates which operations are used in a cell. In general, the GCN-based predictor uses both matrices as input to predict the performance of an architecture, and the MLP-based predictor only uses the operations matrix. For example, Neural Predictor \cite{nrlprdctr} and BRP-NAS \cite{brpnas} built their predictor with GCN. However, WeakNAS \cite{weaknas} just applied a fancy sampling method with a weaker predictor MLP to obtain a significant improvement over BRP-NAS. It is noticed that MLP does not combine prior adjacent trails of operations (adjacency matrix) as GCN does. The fact may indicate that this prior knowledge may not be necessary. It inspires us to explore the gap between GCN and MLP. In our experiments, we found that GCN is only sometimes better than MLP. It is even worse than MLP in many experiment settings.
\begin{figure}[t]
\centering
\includegraphics[width=8.cm]{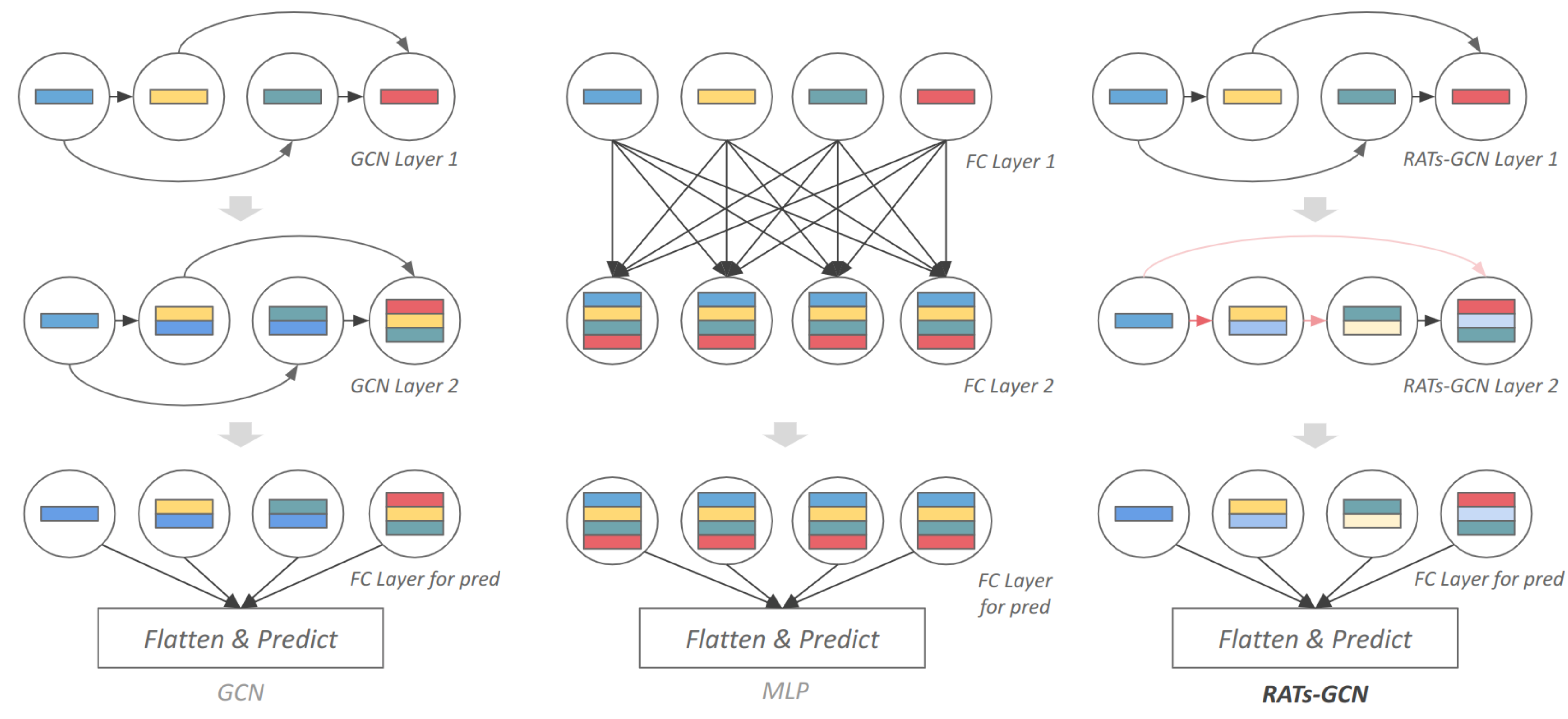}\\
\caption{The illustration of how different predictors transfer features. The four circles in each column represent the four operations in the cell, and the colors in the circles represent the current features of the operation. The trails' thickness and the color's depth represent the weight of 0$\sim$1, respectively.}
\label{concepofrats}
\end{figure}
This phenomenon may be due to the information propagation barrier caused by the inherent adjacent trails and matrix multiplication in GCN. Therefore, the proposed Redirected Adjacent Trails GCN (RATs-GCN) is an adaptive version between GCN and MLP. It has prior knowledge of adjacent trails and avoids the information transmission obstacles that GCN may cause. It can change trails by itself through learning and replace the binary state of trails with weight [0,1]. In addition, based on our observations on cell-based NAS methods, we think architectures with similar FLOPs will perform similarly.  Then, we propose a Predictor-based Search Space Sampling (P3S) module to rapidly focus on the tighter FLOPs intervals of the search space to efficiently search for the desired architecture. Finally, the proposed RATs-NAS method surpasses WeakNAS and Arch-Graph \cite{archnas} and achieves SoTA performance on NASBench-201. 

\section{RELATED WORK}
\subsection{Various Types of NAS}
There have been many studies on NAS in the past. Some methods are based on reinforcement learning \cite{nas, dsgnasrl, mnas}, and others are developed from the evolutionary algorithm \cite{nsganet, reicnas, lsenas, genticcnn}. The predictor-based NAS methods focuse on training a predictor to predict the performance of a CNN architecture and quickly filter out impossible candidates \cite{pnas, nao, nrlprdctr, brpnas, npenas, weaknas, archnas}.  It can reduce the verification time required to evaluate the performance of an architecture candiate.  The cell-based search space shares a fixed meta-architecture and the same hyperparameters. Therefore, the search space is reduced to a small cell which contains several operations such as convolution, pooling, etc. The predictors can be GCN-based or MLP-based. The GCN-based predictor performs more accurately, but needs more training time than the MLP-based one.  This paper proposes a predictor called RATs-GCN which combines both advantages of GCN and MLP to better find the desired architecture.
\subsection{Predictor-based NAS}
There are many types of these predictors \cite{pnas, nao, nrlprdctr, brpnas, npenas} for NAS. In recent work, the Neural Predictor \cite{nrlprdctr} is the most common method and encodes a cell as an adjacency matrix and an operations matrix. The adjacency matrix indicates the adjacent trails of operations and the operations matrix indicates the features of operations. The GCN-based predictor shows significant performance in NASBench-101 \cite{nasbench101}. Since the number of cells in its meta-architecture is equal, the hyperparameters of GCN are fixed. It uses multiple graph convolution \cite{gcn} to extract high-level features and directionality from the above two matrices and then uses a fully connected layer to get the prediction.  After that, the promising architecture is found with the prediction from the search space. Moreover, BRP-NAS \cite{brpnas} proposes a binary predictor that simultaneously takes two different architectures as input and predicts which is better rather than directly predicting their accuracies. This method dramatically improves the prediction performance compared to Neural Predictor \cite{nrlprdctr}. WeakNAS~\cite{weaknas} proposes a more robust sampling method and then adopts MLP to form the predictor. Surprisingly, even though a weak MLP-based preditor is used, it still surpasses Neural Predictor and BRP-NAS. The Arch-Graph \cite{archnas} proposes a transferable predictor and can find promising architectures on NASBench-101 and NASBench-201 \cite{nasbench201} on a limited budget.

\begin{figure}[t]
\centering
\includegraphics[width=7cm]{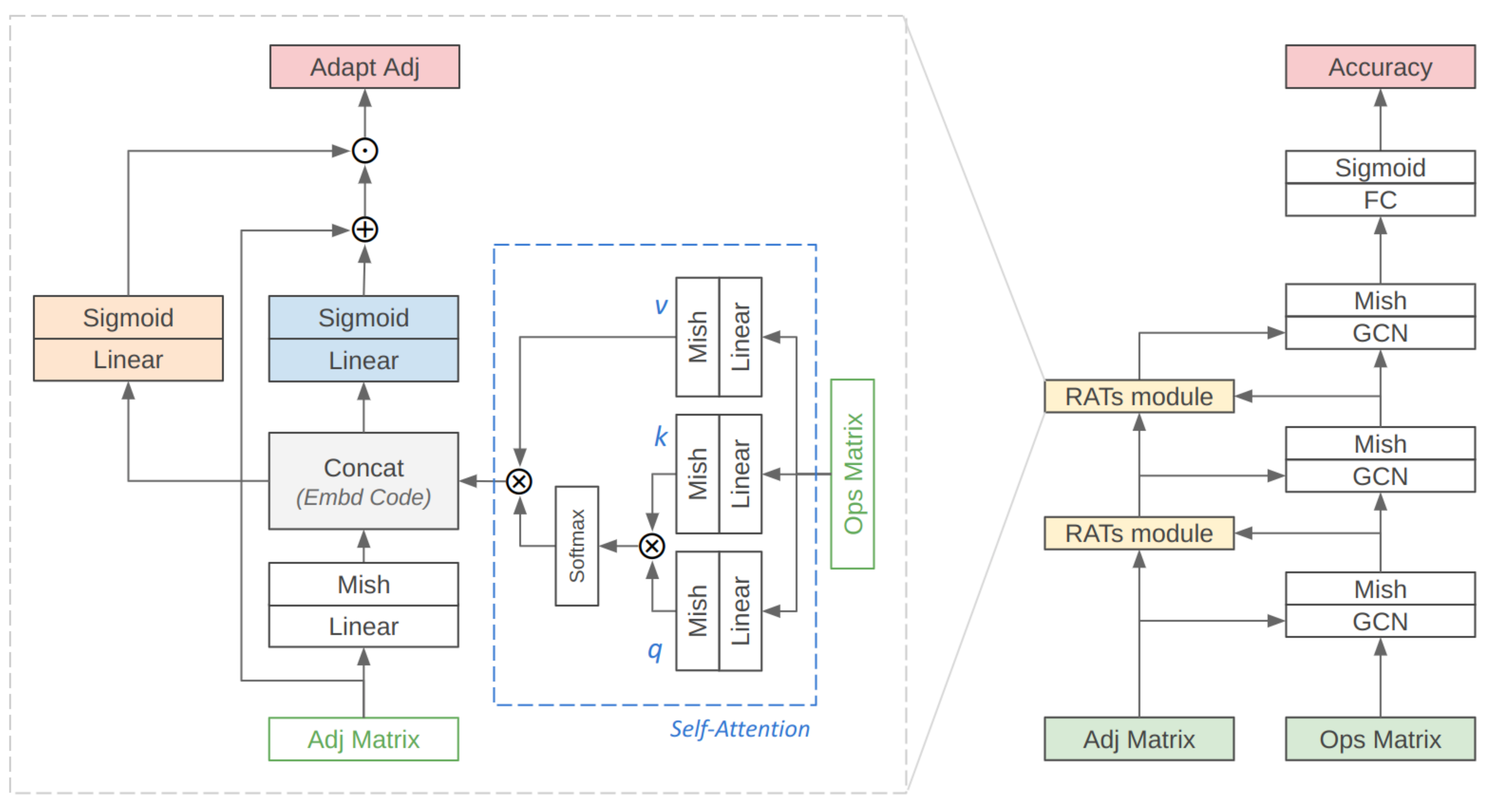}\\
\caption{Architecture of RATs-GCN (right) and the design of RATs module (left). The blue part is used to get offsets, and the orange part is used to get strength. $\bigodot$ denotes Hadamard product, $\bigotimes$ denotes matrix multiplication, and $\bigoplus$ denotes element-wise addition.}
\label{ratsgcnarch}
\end{figure}
\begin{figure}[t]
\centering
\includegraphics[width=7cm]{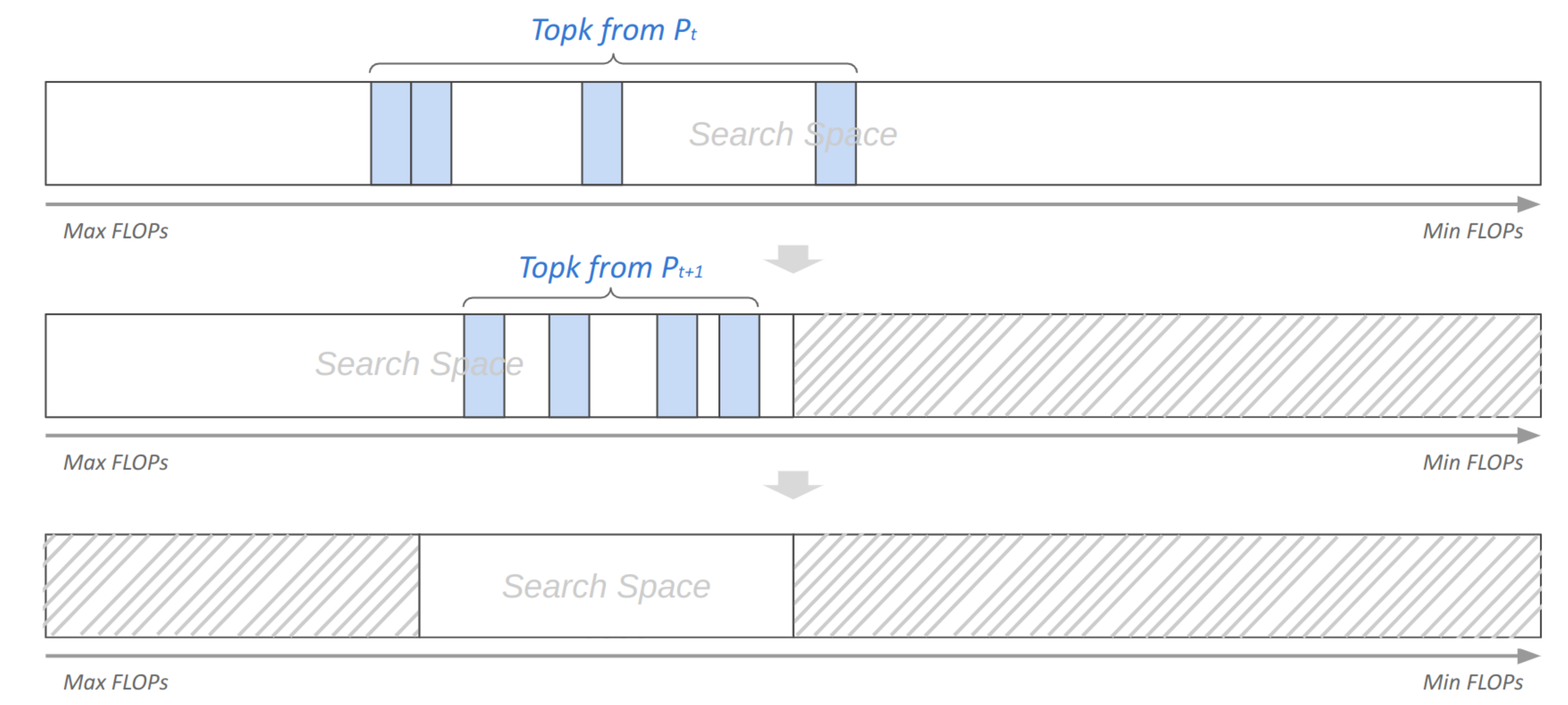}\\
\caption{ The Predictor-based Search Space Sampling (P3S) module reveals an iterative process from top to bottom, which may reduce the attention interval each time, and select top $k$ architectures from this interval using predictor $P$ at time $t$.}
\label{dss}
\end{figure}
\begin{figure*}[ht]
\centering
\begin{tabular}{ccc}
{\includegraphics[width=5.cm]{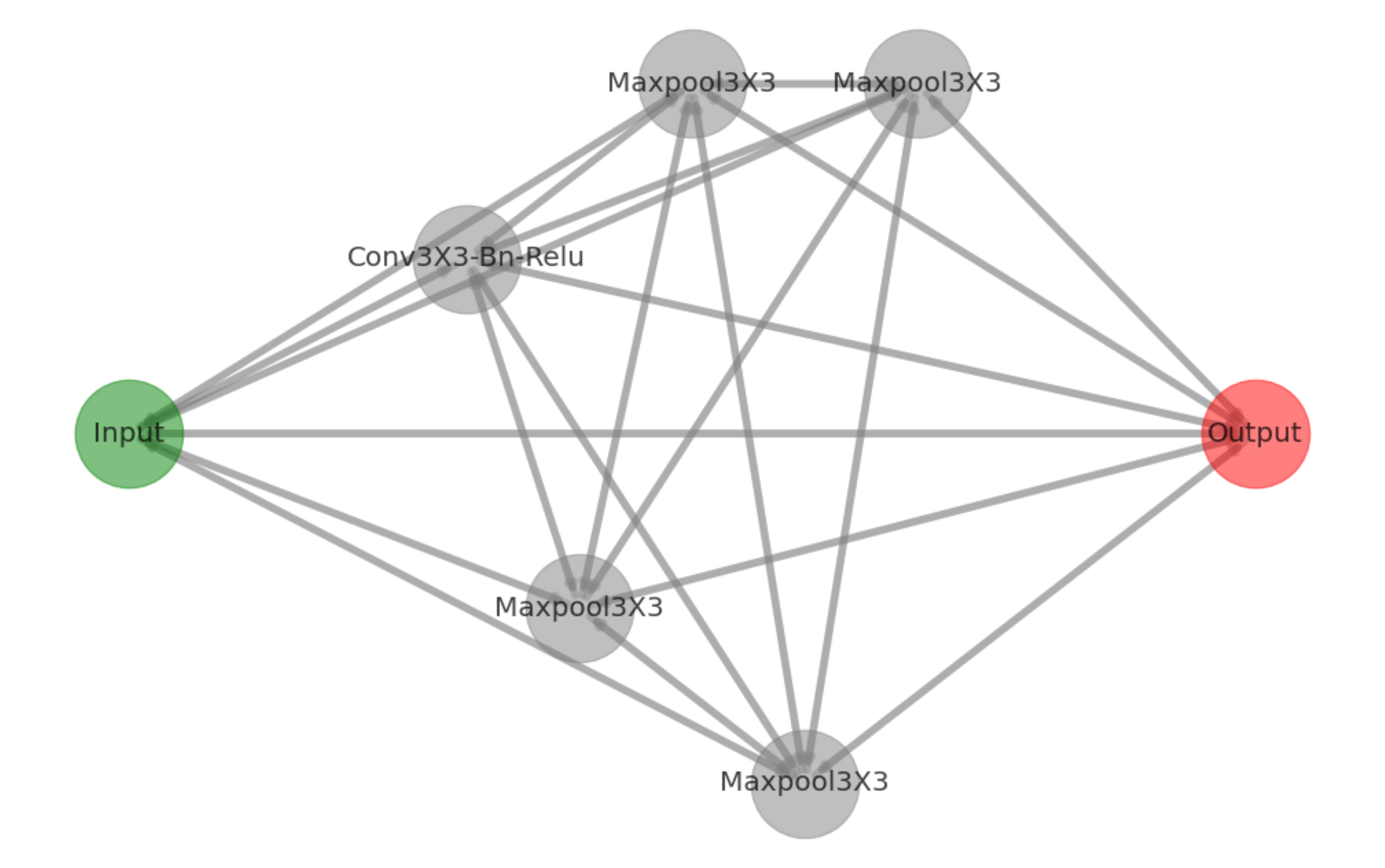}}&
{\includegraphics[width=5.cm]{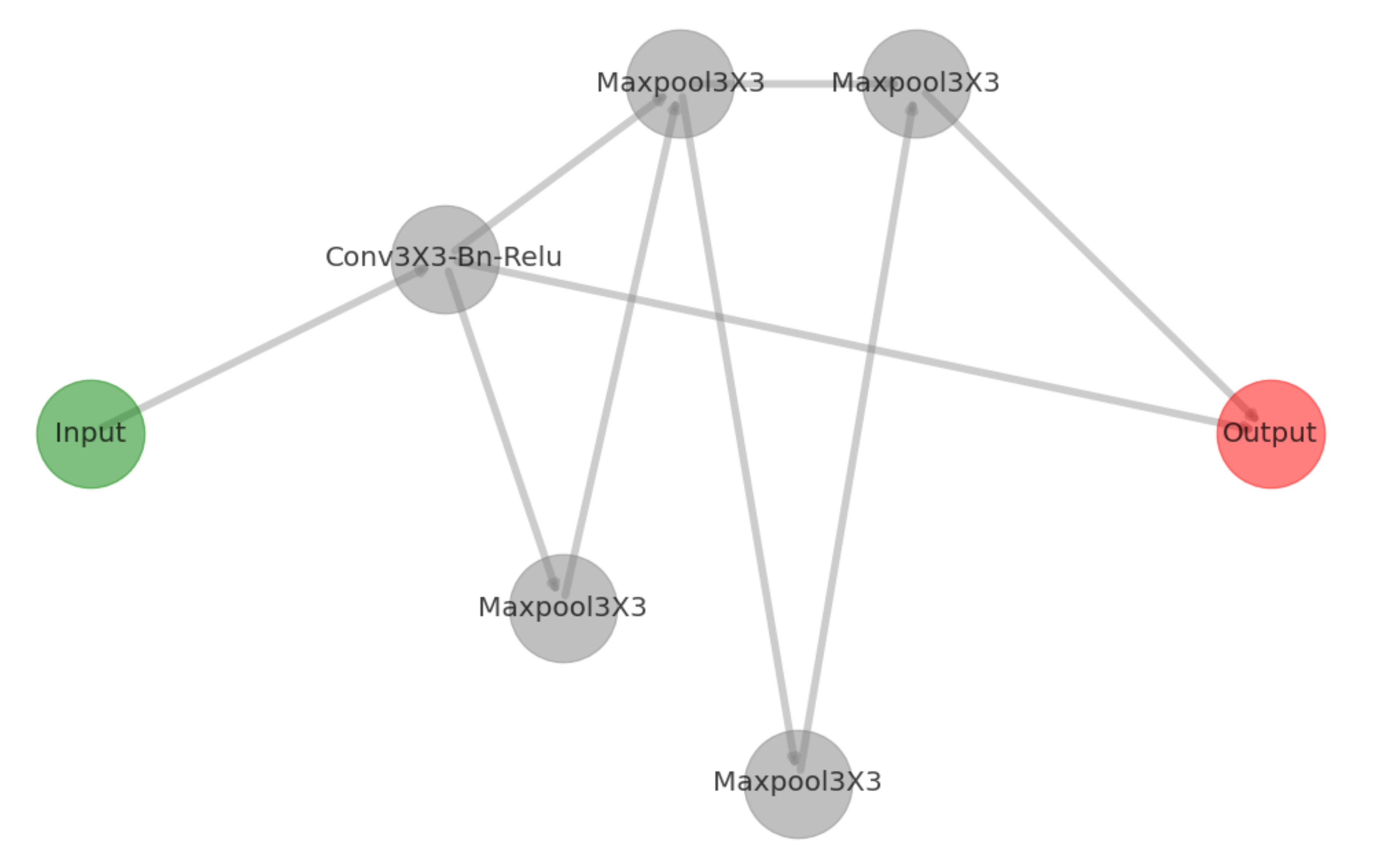}}&
{\includegraphics[width=5.cm]{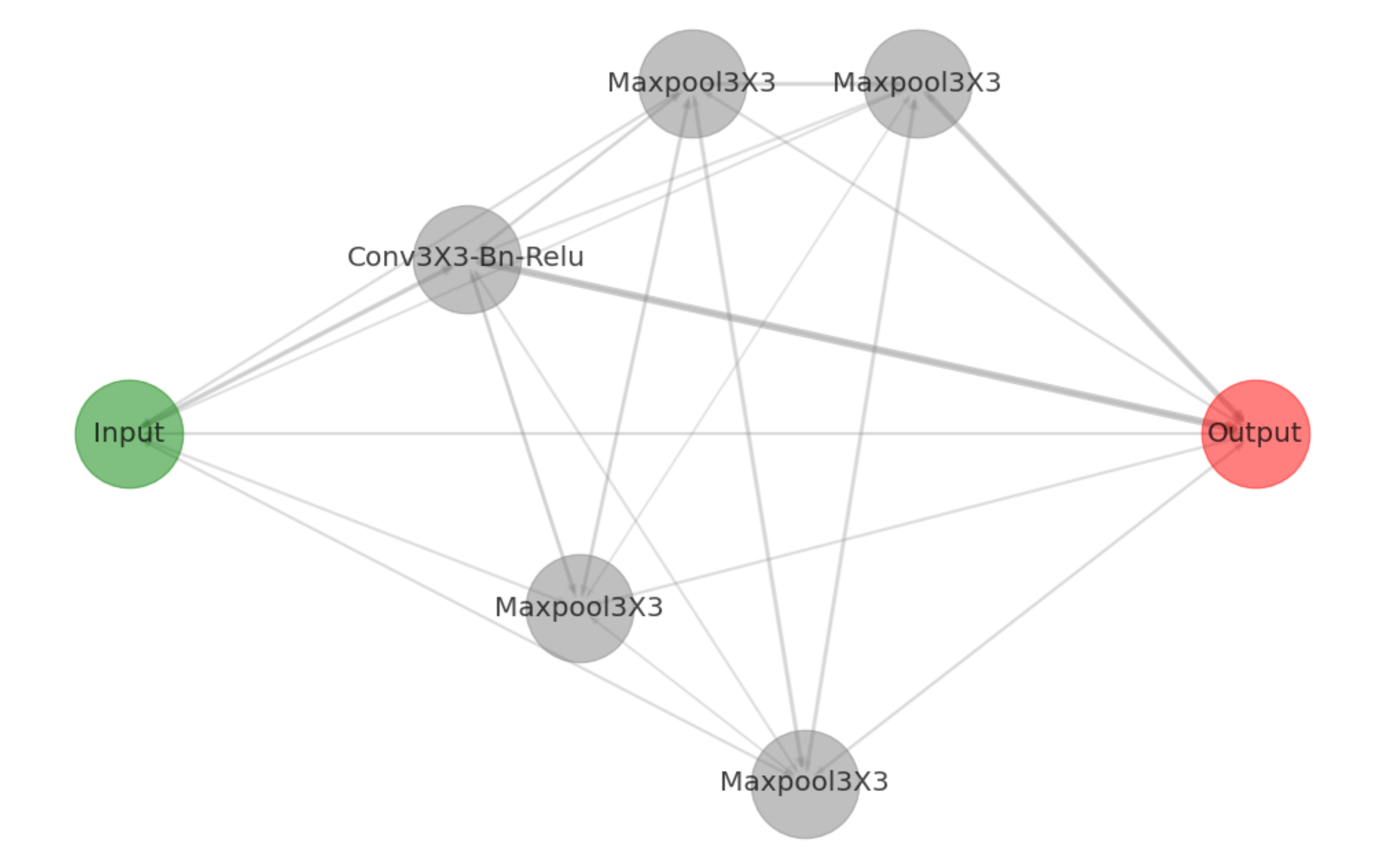}}\\
(a) MLP & (b) GCN & (c) RATs-GCN
\end{tabular}
\caption{The green node represents the input tensor, the red node represents the output tensor, and the other nodes represent respective operations. Directed adjacent trails connect these nodes. This figure is a randomly sampled NASBench-101 cell, and we plot the adjacent trails under three types of predictors: (a) MLP, (b) GCN, and (c) RATs-GCN.}
\label{visulizstionofrats}
\vspace{-1.0em}
\end{figure*}
\section{APPROACH}
\subsection{Redirected Adjacent Trails GCN (RATs-GCN)}
As mentioned above, the predictors can be GCN-based or MLP-based. As shown in \myfigref{concepofrats}, GCN uses adjacent trails of operations (adjacency matrix) and operation features (operation matrix) in a cell as input. MLP uses only operations features as input, which means that GCN has more prior knowledge than MLP, and MLP can be regarded as a full network connection architecture of adjacent trails. However, as shown in Tab. \ref{cmpforgcns}, we found that GCN is only sometimes better than MLP in all experimental settings since its inherent adjacent trails hinder the information flow caused by matrix multiplication. The adjacency matrix used in GCN may receieve the negative effects caused by the directions stored in the inherent adjacent trails. To address this problem, the trails stored in the adjacency matrix should be adaptively changed. Thus, this paper proposes a RATs (Redirected Adjacent Trails) module and attachs it to the backbone of GCN to adaptively tune the trail directions and their weights stored in the adjacency matrix. This module allows GCN to change each trail with a new learning weight. In extreme cases, RATs-GCN can be GCN or MLP.

\subsection{Redirected Adjacent Trails Module (RATs)}
As described above, the trails stored in the adjacency matrix of GCN are fixed.  If the trails are wrongly set, negative effects will be sent to the predictor and result in accuracy deficiency. Fig. \ref{concepofrats} shows the concept of our RATs predictor. The trails in the original GCN-based predcitor are fixed (the left part of Fig. \ref{concepofrats}) during training and inference.  However, in the right part of Fig. \ref{concepofrats},  the trails in our RATs-GCN will be adaptively adjusted according to the embedded code generated by the operation matrix. Fig. \ref{ratsgcnarch} shows the detailed designs of the RATs module and RATs-GCN. Unlike the original GCN-based predictor, a new RATs module is proposed and attached to our RATs-GCN predictor.  This module first converts the operation matrix to three new feature vectors: query $Q$, key $K$, and vaule $V$ by self-attention \cite{vit}. Then, an embedded code can be obtained by concatenating $Q$, $K$, $V$ with the original adjacency matrix. With this embedded code as input, the trail offsets and operation strengths are generated by a linera projection and sigmoid function.  Finally, a new adjacency matrix is generated by adding the offset to the adjacency matrix and then doing a Hadamard product with the strength. The RATs can redetermine the trails and their strengths.

\begin{table}[t]
\caption{The comparison of RATs-GCN and the other predictors on NASBench-101 and NASBench-201. The mACC is the mean accuracy of the top 100 architectures ranked from the predictor. The Psp is Spearman Correlation, and the calculation range is the entire search space.}
\begin{center}
\resizebox{210pt}{120pt}{
\begin{tabular}{c|c|c|c|c|c|c|c}
\hline
\multicolumn{8}{c}{NASBench-101 (423,624 unique cells.)}\\
\hline
& Budgets& \multicolumn{3}{|c|}{mAcc (\%)} & \multicolumn{3}{|c}{Psp (\%)}\\
\hline
MLP & 300 & \multicolumn{3}{|c|}{90.78} & \multicolumn{3}{|c}{30.38}\\
GCN & 300 & \multicolumn{3}{|c|}{89.54} & \multicolumn{3}{|c}{1.93}\\
BI-GCN & 300 & \multicolumn{3}{|c|}{91.48} & \multicolumn{3}{|c}{43.82}\\
\textbf{RATs-GCN} & 300 & \multicolumn{3}{|c|}{\textbf{92.80}} & \multicolumn{3}{|c}{\textbf{60.80}}\\
\hline
MLP & 600 & \multicolumn{3}{|c|}{91.72} & \multicolumn{3}{|c}{42.87}\\
GCN & 600 & \multicolumn{3}{|c|}{91.04} & \multicolumn{3}{|c}{18.52}\\
BI-GCN & 600 & \multicolumn{3}{|c|}{91.56} & \multicolumn{3}{|c}{38.56}\\
\textbf{RATs-GCN} & 600 & \multicolumn{3}{|c|}{\textbf{92.94}} & \multicolumn{3}{|c}{\textbf{70.24}}\\
\hline
MLP & 900 & \multicolumn{3}{|c|}{92.03} & \multicolumn{3}{|c}{48.45}\\
GCN & 900 & \multicolumn{3}{|c|}{90.94} & \multicolumn{3}{|c}{27.16}\\
BI-GCN & 900 & \multicolumn{3}{|c|}{92.15} & \multicolumn{3}{|c}{53.71}\\
\textbf{RATs-GCN} & 900 & \multicolumn{3}{|c|}{\textbf{93.01}} & \multicolumn{3}{|c}{\textbf{70.58}}\\
\hline
\multicolumn{8}{c}{NASBench-201 (15,625 unique cells.)}\\
\hline
& \multirow{2}*{Budgets} & \multicolumn{2}{|c|}{CIFAR-10} & \multicolumn{2}{|c|}{CIFAR-100} & \multicolumn{2}{|c}{ImgNet-16}\\
\cline{3-8}
& ~ & mAcc & Psp & mAcc & Psp & mAcc & Psp\\
\hline
MLP & 30 & 88.54 & 10.39 & 64.68 & 19.39 & 37.98 & 22.97\\
GCN & 30 & 84.86 & -0.04 & 65.12 & 31.89 & 37.69 & 33.06\\
BI-GCN & 30 & 86.26 & 21.02 & 61.96 & 34.06 & 38.28 & 40.61\\
\textbf{RATs-GCN} & 30 & \textbf{89.68} & \textbf{47.61} & \textbf{69.81} & \textbf{65.72} & \textbf{43.11} & \textbf{67.18}\\
\hline
MLP & 60 & 90.93 & 27.36 & 68.31 & 42.68 & 41.49 & 47.32\\
GCN & 60 & 87.87 & 29.93 & 67.42 & 42.77 & 38.89 & 44.05\\
BI-GCN & 60 & 87.82 & 18.24 & 64.28 & 47.39 & 39.32 & 54.22\\
\textbf{RATs-GCN} & 60 & \textbf{92.72} & \textbf{61.67} & \textbf{70.05} & \textbf{73.60} & \textbf{43.79} & \textbf{74.64}\\
\hline
MLP & 90 & 91.69 & 42.86 & 65.64 & 51.72 & 41.98 & 56.22\\
GCN & 90 & 90.83 & 40.30 & 67.64 & 44.99 & 39.15 & 45.51\\
BI-GCN & 90 & 89.71 & 36.14 & 68.11 & 62.22 & 42.17 & 65.51\\
\textbf{RATs-GCN} & 90 & \textbf{93.17} & \textbf{70.50} & \textbf{69.66} & \textbf{74.98} & \textbf{44.16} & \textbf{77.39}\\
\hline
\end{tabular}}
\end{center}
\label{cmpforgcns}
\vspace{-2.0em}
\end{table}

\subsection{Predictor-based Search Space Sampling (P3S)}
Although the proposed RATs-GCN has already provided more flexible plasticity than GCN and MLP, we all know that a predictor-based NAS's performance depends not only on predictor design but also on the sampling method. WeakNAS gets SOTA performance using a weaker predictor with a firmer sampling method. So, a promising strategy is bound to bring about considerable improvement for NAS. The proposed P3S is based on our observations on cell-based NAS: (1) The architectures constructed in a cell-based approach share the same meta-architecture and candidate operations for cell search; (2) Each layer in the meta-architecture has the same hyperparameters, such as filters, strides, etc.  This means that those candidate operations for cells have the same input and output shapes. In short, there are many architectures that are very similar because they all share the same meta-architecture and limited candidate operations, and the hyperparameters are the same. All of them result in our P3S method. The P3S method rapidly divides search space into tighter FLOPs intervals by following rough steps: (1) Sort the search space $S$ by FLOPs and initialize $i_1=0$ and $i_2=len(S)-1$ as the focus interval; (2) Select the top 1\% architectures of the sub search space $[S_i, S_j]$ sorted by predictor at $t$ time $P_t$; (3) If the indexes of selected architectures are at least 75\% in the first or second half of the search space sorted by FLOPs, move $i, j$ to that half interval; if not, move $i, j$ to the last interval; (4) Sorting and get the top $k$ of $[S_i, S_j]$ by $P_t$ and add these $S_{topk}$ to the sample pool $B$; (5) Training $P_{t+1}$ based on $B$, then back to (2). At the beginning of $t=0$, we randomly select $k$ samples from the search space as initial training samples to train $P_0$. After that, the above steps will continue until we find the optimal global cell or exceed the budget. This process aims to rapidly divide and focus on the tighter FLOPs range because we believe the architectures with similar FLOPs will perform similarly. P3S has corrective measures such as Step (2) to avoid falling into the wrong range.

\begin{table}[t]
\caption{The comparison on the number of samples required to find the global optimal on NASBench-201.}
\begin{center}
\resizebox{200pt}{42pt}{
\begin{tabular}{c|c|c|c}
\hline
&\multicolumn{3}{c}{NASBench-201}\\
\cline{2-4}
& CIFAR-10 & CIFAR-100 & ImgNet-16\\
\hline
Random Search & 7782.1 & 7621.2 & 7726.1\\
Reg Evolution & 563.2 & 438.2 & 715.1\\
MCTS & 528.3 & 405.4 & 578.2\\
LaNAS & 247.1 & 187.5 & 292.4\\
WeakNAS & 182.1 & 78.4 & 268.4\\
\textbf{RATs-NAS} & \textbf{114.6} & \textbf{74.3} & \textbf{146.7}\\
\hline
\end{tabular}}
\end{center}
\label{findoptimal}
\vspace{-2.0em}
\end{table}

\section{EXPERIMENTS}
\subsection{Comparison of RATs-GCN and GCNs and MLP}
We extensively test MLP, GCN, BI-GCN, and RATs-GCN under similar model settings with 30 runs for a fair comparison. GCN, BI-GCN, and RATs-GCN have three GCN layers with 32 filters and one FC layer with one filter to obtain output prediction accuracy, and MLP has three FC layers with 32 filters and one FC layer with one filter. They all applied the random sampling method to get training architectures from the search space. We evaluated them on NASBench-101 with training budgets of 300, 600, and 900. We also evaluate them on the three sub-datasets (CIFAR-10, CIFAR-100, ImageNet-16) of NASBench-201 with training budgets 30, 60, and 90. As shown in \mytabref{cmpforgcns}, we found that RATs-GCN surpasses others in mAcc for about 1\%$\sim$5\% and in Psp for about 10\%$\sim$50\% under different budgets. The mAcc denotes the average accuracy of the top 100 architectures predicted by the predictor. The Psp denotes the Spearman Correlation of predicted ranking and ground truth ranking.

\subsection{Comparison of RATs-NAS and SOTAs}
We design two experiments with 30 runs to verify the performance of RATs-NAS and compare it with other SOTA methods. The first experiment aims to evaluate how fast an NAS method finds the optimal one in the search space. As shown in \mytabref{findoptimal}, we can see that the RATs-NAS use fewer architectures than other methods. It even finds the global optimal cell using an average of 146.7 architecture costs, nearly twice as fast as WeakNAS. Another experiment examines how good architecture can be found at the cost of 150 architectures. As shown in \mytabref{150best}, the RATs-NAS find the architecture with an accuracy of 73.50\% on NASBench-201 (CIFAR-10), it better than other SOTA methods. Considering the optimal accuracy are 94.37\%, 73.51\%, 47.31\% in three sub-dataset of NASBench-201, it can find the architectures of 94.36\%, 73.50\%, 47.07\% with such little cost. It shows a significant performance and beats others by a considerable margin.

\subsection{Visualization of Adjacent Trails}
In order to obtain more evidence to support the RATs module, in addition to other experiments focusing on performance, we also visualize the trail of operations in a single cell. We randomly select an architecture (cell) from NASBench-101, then draw its adjacent trails to represent GCN, draw full trails to represent MLP, and draw new adjacent trails by the last RATs module in RATs-GCN. As shown in Fig. \ref{visulizstionofrats}, we can see that the proposed RATs-NAS differs from GCN and MLP. Part (c) of \myfigref{visulizstionofrats} shows that RATs-GCN gets approximate MLP trails with weights between 0 and 1 starting from GCN trails.

\begin{table}[t]
\caption{The comparison of RATs-NAS and SOTAs on NASBench-201. Note that the methods NP- are based on \cite{nrlprdctr} and we replace its predictor with several types.}
\begin{center}
\resizebox{200pt}{60pt}{
\begin{tabular}{c|c|c|c}
\hline
&\multicolumn{3}{c}{NASBench-201}\\
\cline{2-4}
& CIFAR-10 & CIFAR-100 & ImgNet-16\\
\hline
NP-MLP & 93.95 & 72.15 & 46.30\\
NP-GCN & 94.04 & 72.37 & 46.28\\
NP-BI-GCN & 94.07 & 72.18 & 46.39\\
\textbf{NP-RATs} & \textbf{94.17} & \textbf{72.78} & \textbf{46.58}\\
\hline
Random Search & 93.91 & 71.80 & 46.03\\
Reg Evolution & - & 72.70 & -\\
BONAS & - & 72.84 & -\\
WeakNAS & 94.23 & 73.42 & 46.79\\
Arch-Graph & - & 73.38 & -\\
\textbf{RATs-NAS} & \textbf{94.36} & \textbf{73.50} & \textbf{47.07}\\
\hline
\end{tabular}}
\end{center}
\label{150best}
\vspace{-2.0em}
\end{table}
\section{Conclusion}
A RATs-GCN predictor was proposed to improve the performance of GCN-based predictors. It can change trails and give the trails different weights and performs much better than GCN, MLP, and BI-GCN on NASBench-101 and NASBench-201. Then we propose the P3S method to rapidly divide the search space and focus on tighter FLOPs intervals. Finally, the proposed RATs-NAS consists of RATs-GCN and P3S outperforms WeakNAS and Arch-NAS on NASBench-201 by a considerable gap.

\vfill\pagebreak
\bibliographystyle{IEEEbib}
\bibliography{refs}

\end{document}